# Revolutionizing Healthcare Image Analysis in Pandemic-Based Fog-Cloud Computing Architectures

Al Zahraa Elsayed[1,*]    Khalil Mohamed[2]    Hany Harb[3]

**Abstract** The emergence of pandemics has significantly emphasized the need for effective solutions in healthcare data analysis. One particular challenge in this domain is the manual examination of medical images, such as X-rays and CT scans. This process is time-consuming and involves the logistical complexities of transferring these images to centralized cloud computing servers. Additionally, the speed and accuracy of image analysis are vital for efficient healthcare image management. This research paper introduces an innovative healthcare architecture that tackles the challenges of analysis efficiency and accuracy by harnessing the capabilities of Artificial Intelligence (AI). Specifically, the proposed architecture utilizes fog computing and presents a modified Convolutional Neural Network (CNN) designed specifically for image analysis. Different architectures of CNN layers are thoroughly explored and evaluated to optimize overall performance. To demonstrate the effectiveness of the proposed approach, a dataset of X-ray images is utilized for analysis and evaluation. Comparative assessments are conducted against recent models such as VGG16, VGG19, MobileNet, and related research papers. Notably, the proposed approach achieves an exceptional accuracy rate of 99.88% in classifying normal cases, accompanied by a validation rate of 96.5%, precision and recall rates of 100%, and an F1 score of 100%. These results highlight the immense potential of fog computing and modified CNNs in revolutionizing healthcare image analysis and diagnosis, not only during pandemics but also in the future. By leveraging these technologies, healthcare professionals can enhance the efficiency and accuracy of medical image analysis, leading to improved patient care and outcomes.

Keywords COVID-19, Deep Learning, Convolution Neural Network, Fog Computing, Healthcare System.

## 1 Introduction

The field of healthcare data analysis has undergone a significant transformation, particularly in response to pandemics, which have heightened the demand for efficient solutions. During such critical times, healthcare professionals and researchers face a significant challenge in manually analyzing medical images, including X-rays and CT scans [1]. This task is time-consuming and complicated by the logistical hurdles of transferring these large image datasets to centralized cloud computing servers. Moreover, the speed and accuracy of image analysis are crucial factors in effective healthcare image management.

Cloud computing, a technology that enables users to access computing resources such as data storage and processing power via the internet, has the potential to enhance the safety, quality, and efficiency of healthcare. One application of cloud computing in healthcare involves storing and analyzing extensive patient data to identify trends and patterns that can aid in more effective disease diagnosis and treatment [1]. Additionally, cloud computing can facilitate the remote delivery of healthcare services, particularly benefiting patients residing in rural areas or those facing challenges in visiting a doctor's office [2]. However, the traditional cloud computing architecture for medical and healthcare purposes relies on a centralized approach for data transmission, which poses several challenges [3]:

- **Security**: Centralized data storage increases vulnerability to cyberattacks.
- **Scalability**: Scaling centralized data storage to meet the needs of a growing number of users can be challenging.
- **Compliance**: Healthcare organizations must adhere to strict regulations governing the privacy and security of patient data, which can be difficult with a centralized storage system.
- **Latency**: The time it takes for data to travel from the network's edge to the cloud and back may be too long for time-sensitive healthcare applications, particularly those requiring rapid response in emergency situations.
- **Cost**: Transferring large amounts of data to the cloud can be expensive.

These challenges have led some healthcare organizations to adopt fog computing as an alternative to cloud computing. Fog computing, a distributed cloud computing architecture, is better suited for healthcare applications. Fog computing nodes are positioned closer to the network's edge, reducing latency and improving security. Moreover, fog computing can be a more cost-effective solution for healthcare applications that involve significant data transfer. Fog computing holds great promise for healthcare and is expected to witness increased adoption in the future [4].

Deep Learning, a subset of artificial intelligence, is being leveraged to enhance healthcare in various ways. Deep learning algorithms can analyze medical images and data to diagnose diseases, develop new treatments, and deliver personalized care [5, 6]. It is also instrumental in developing healthcare applications such as virtual assistants and chatbots. Deep Learning is a rapidly evolving field that is poised to have a significant impact on healthcare in the years to come [7, 8].

[1,2,3]Systems and Computers Engineering Dept., Faculty of Engineering, Al-Azhar University, Nasr city, Cairo, Egypt. Corresponding author Email: eng.khalil@azhar.edu.eg

To address these formidable challenges, this research paper introduces an innovative healthcare architecture that aims to achieve both analysis efficiency and accuracy. At its core, this architecture harnesses the power of Artificial Intelligence (AI). Specifically, it presents a novel framework that combines the fog computing paradigm with a meticulously designed modification of Convolutional Neural Networks (CNNs), tailored explicitly for image analysis. A comprehensive exploration of various CNN layer architectures is conducted and subjected to rigorous evaluation to optimize performance.

To empirically validate the effectiveness of the proposed approach, a curated dataset of COVID-19-related X-ray images is utilized for analysis and evaluation. These images serve as a practical testbed to enable comparative assessments against contemporary models such as VGG16, VGG19, and MobileNet. The outcomes of these assessments unequivocally demonstrate the exceptional potential of the proposed framework.

These compelling results underscore the transformative potential that emerges from the intersection of fog computing and modified CNNs in the domain of healthcare image analysis. In a world grappling with the challenges posed by pandemics, the convergence of cutting-edge technology and medical science holds the promise of revolutionizing healthcare image analysis and diagnosis, not only during times of crisis but also in the broader landscape of healthcare delivery. This research represents a critical step forward in realizing this promise and provides a glimpse into a future where the fusion of AI and healthcare unlocks boundless potential.

The remainder of this paper is organized as follows: Section 2 discusses previous work in the field. Section 3 describes the proposed framework. Section 4 presents the convolutional neural networks (CNNs) architecture. Section 5 introduces mathematical formulas for the proposed model. Section 6 discusses database for Chest X-rays. Section 7 discusses the experimental results. Section 8 evaluation of two and three Layers of the CNN Model with different Epochs Finally, Section 9 concludes the paper by summarizing the findings and outlining future research directions.

## 2 Related Works

The advancement of science and technology has historically been driven by medicine and healthcare [9, 10]. Recent research has focused on integrating fog computing into Internet of Health Technology (IoHT) applications, yielding positive outcomes such as reduced service response time, improved system performance, and increased energy efficiency. For instance, Xue et al. [11] developed the analytic network technique to identify and rank fog computing-based IoT solutions for health system monitoring. Fog computing in healthcare involves establishing a distributed intermediate layer between the cloud and sensor hubs using IoT frameworks.

Gia et al. [12] demonstrated the use of fog computing as a gateway to enhance health monitoring systems. They created fog computing features, including interoperability, a distributed database, a real-time notification mechanism, position awareness, and a graphical user interface with access management. Additionally, they presented a lightweight, customizable framework for extracting ECG features (e.g., heart rate, P wave, and T wave).

Elhadad et al. [13] proposed a fog-based health monitoring framework that utilizes fog gateways for clinical decision-making based on data collected from sensors embedded in wearable devices. These sensors, such as temperature sensors, ECG sensors, and blood pressure sensors, measure a patient's temperature, heart rate, and systolic and diastolic pressure.

Al-Khafajiy et al. [14] introduced the concept of IoT-fog computing in IoT-based healthcare systems, suggesting a methodology for improving fog performance through collaborative policies among fog nodes for optimal workload and job distribution. Similarly, El-Rashidy et al. [15] presented a comprehensive strategy for monitoring pregnant women, utilizing a Data Replacement and Prediction Framework (DRPF) divided into three layers: (i) IoT, (ii) Fog, and (iii) Cloud. Their findings indicated strong associations between patient age, BMI, blood pressure, lymphocyte vitamin E levels, and the diagnosis of gestational diabetes mellitus (GDM).

Quy et al. [16] presented an all-in-one computer architectural framework and conducted a survey of IoT applications based on fog computing in the healthcare industry. They explored the application potential, challenges, and future research objectives in this field. Similarly, Shi et al. [17] discussed the vision and essential characteristics of fog computing, which aims to address the latency issue caused by IoT by distributing processing, storage, and networking resources to the edge of the network, interacting with the cloud.

Arunkumar et al. [18] proposed HealthFog-CCNN, a fog-based Smart Healthcare System for Automatic Diagnosis of Heart Diseases that combines deep learning and IoT. Their research focused on the medical aspects of heart disease patients, utilizing deep learning in edge computing devices for real-time analysis of heart problems.

Lastly, Mutlag et al. [19] aimed to contribute to the existing knowledge by providing specific examples categorized into four groups: fog computing methods in healthcare applications, system development in fog computing for healthcare applications, and evaluation and surveys of fog computing in healthcare applications.

While fog computing and artificial intelligence have been effectively utilized in the healthcare field, no effective framework has been used for heavy healthcare processing, particularly during a pandemic. Furthermore, most papers examine small datasets for efficiency testing. Therefore, this paper proposes a novel framework based on a distributed data ingestion/collection layer, with distributed data processing and integration layers to simplify the processing of similar data on specific devices. The collected data is then forwarded to fog nodes for

## 3 The Proposed Framework

The integration of fog computing, cloud resources, and artificial intelligence (AI) technologies in the proposed healthcare framework creates a comprehensive and efficient healthcare ecosystem [9, 10] as depicted in Fig. 1. The framework aims to improve patient care, streamline healthcare processes, and support research and innovation. It consists of several interconnected layers, each with its unique roles and functionalities.

- **The first layer is the Data Ingestion Layer**, which plays a crucial role in collecting and assimilating diverse data from sources such as medical devices, sensors, electronic health records, and patient-generated information. Its main functions include data collection, transformation, quality assurance, and aggregation. This layer ensures data compatibility, reliability, and security through encryption and authentication. It also prioritizes scalability, reliability, adherence to healthcare regulations, and interoperability for efficient data sharing and collaboration [11, 12].

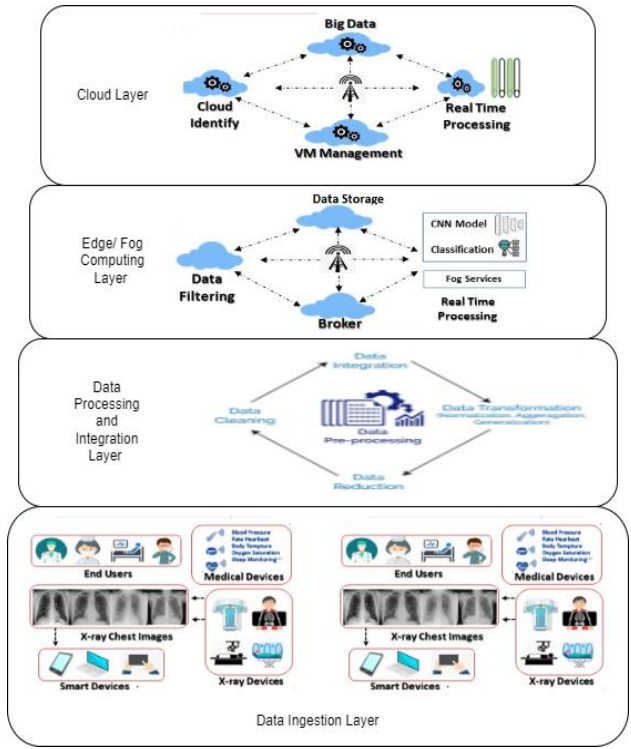

**Fig. 1**. The framework based on IoT fog-cloud computing architecture

- **The second layer is the Data Processing and Integration Layer**, which acts as an intermediary between the data ingestion and advanced analytics layers. Its primary responsibility is to transform and standardize raw healthcare data from different sources, ensuring a uniform structure and format. This layer performs data normalization, quality assurance, aggregation, and real-time processing to generate accurate insights and support timely decision-making. It also maintains interoperability, scalability, data governance, compliance standards, and data integrity, laying the foundation for data-driven healthcare practices and improved patient outcomes [13, 14].
- **The third layer is the Edge/Fog Computing Layer**, which serves as a critical bridge between the data sources and the centralized cloud infrastructure. Positioned closer to the data sources, this layer enables real-time or near-real-time data processing, particularly important for low-latency healthcare applications like patient monitoring and emergency care. It leverages distributed edge or fog computing nodes to execute data analytics and computational tasks at the source, reducing the burden on central cloud resources and optimizing bandwidth usage. This layer ensures local data processing, secure data storage, and efficient resource allocation to enhance system scalability, reliability, and rapid decision-making [15-17].
- **The fourth layer is the Cloud Layer**, which serves as the central component of the healthcare system, providing hardware resources and high-capacity computer services as data centers for data computation and storage. It encompasses data analysis and pre-processing procedures, supporting medical professionals in making long-term treatment decisions. This layer involves various processing tasks, including normalization and data preparation, before training machine learning algorithms such as Convolutional Neural Networks (CNN) for disease diagnosis, predictive modeling, anomaly detection, and data-driven decision support. It contributes to improved healthcare outcomes and drives innovation within the healthcare ecosystem [18, 19].

In summary, the proposed healthcare framework integrates fog computing, cloud resources, and AI technologies to create a holistic and efficient healthcare ecosystem. The framework consists of the Data Ingestion Layer, Data Processing and Integration Layer, Edge/Fog Computing Layer, and Cloud Layer, each playing a vital role in collecting, processing, analyzing, and storing healthcare data to support improved patient care and decision-making.

## 4 The CNN Architecture

Like other domains, the healthcare field has successfully implemented several deep learning (DL) applications that have shown remarkable results in various medical scenarios. This success can be attributed to two main factors: (i) the ability of DL models to learn from labeled or unlabeled datasets and (ii) the inherent risk of human error in diagnosing cases, regardless of the doctors' expertise level. Consequently, the medical research community has developed numerous healthcare systems based on DL methods.

One specific DL technique, called Convolutional Neural Network (CNN), has been designed to excel in image identification, classification, and prediction tasks. CNN's ability to automatically extract features from images and perform in-depth analysis makes it ideally suited for training in the proposed architecture. In this study, a CNN-based model is proposed for detecting chest diseases in patients using chest radiography images. The model's objective is to classify and identify chest diseases by distinguishing between normal X-rays and abnormal chest X-rays, as depicted in Fig. 2

For CNN to operate effectively, the input images must undergo processing to extract visual patterns. This process involves a linear operation where two functions represented by matrices are multiplied to produce an output. Initially, the images are transformed into a matrix format to facilitate information extraction.

CNN comprises several layers that work together to perform these operations efficiently, including (see Fig. 3):

- **Input data layer**: This layer reads a pre-processed collection of images. In our case, the X-ray and CT scan images are pre-processed separately.
- **Convolutional layer**: Serving as the core of our proposed model, this layer is responsible for extracting features from the image collection while preserving the spatial relationship between pixels.
- **Batch Normalization Layer**: This layer is a crucial training strategy in deep neural networks as it ensures the stability and proper training of convolutional features.
- **ReLU layer**: This layer replaces negative pixel values in the convolved features with zero, generating a non-linearity map of the CNN network's features.
- **Fully Connected Layer**: This layer categorizes the convolved features from the image datasets into the desired classes.
- **Softmax Layer**: Interpreting the probability values of the activation function from the previous layer, this layer is particularly relevant for illness diagnosis. The results can be interpreted as two classes: '0' for negative (normal chest X-ray or CT) and '1' for positive disease.

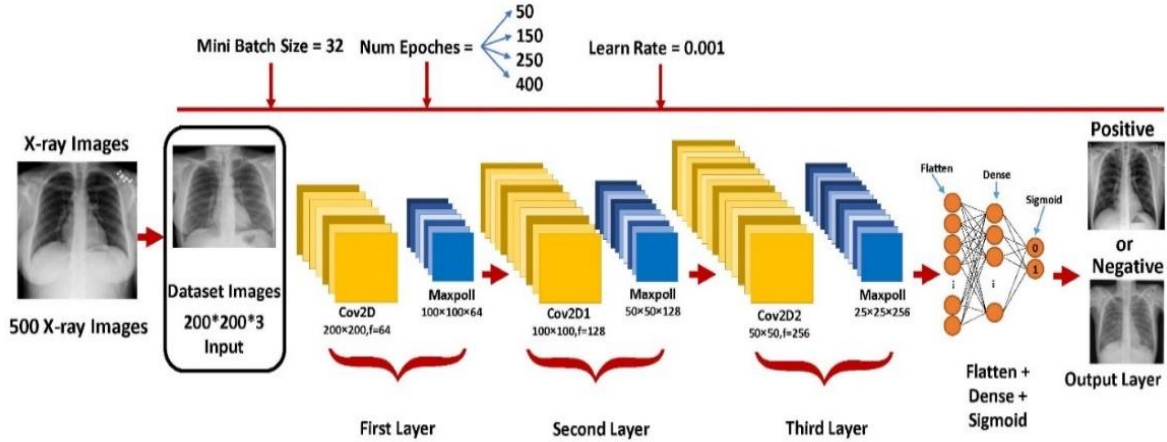

**Fig. 2**. The proposed model is based on CNN architecture.

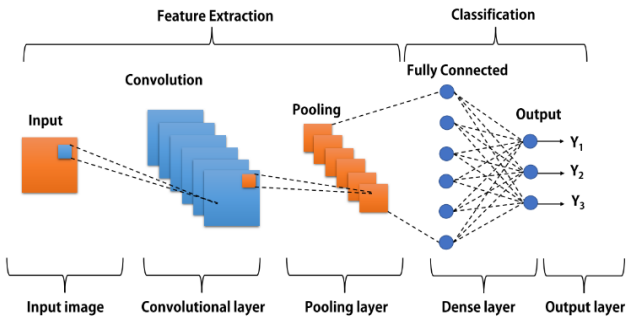

**Fig. 3**. The CNN Architecture layers

- **Output Layer**: This final layer of the CNN model labels the results obtained from the previous layer accordingly [20].

In conclusion, the healthcare field has embraced deep learning applications, leveraging the remarkable capabilities of DL models in various medical scenarios. The proposed architecture incorporates CNN, a specialized DL technique for image identification, classification, and prediction, to detect chest diseases from chest radiography images. The CNN model consists of multiple layers, each playing a distinct role in efficiently processing and analyzing the input data to classify and identify chest diseases [20].

## 5 Mathematical Formulas

In this paper, to evaluate our model, the cross-entropy log loss function between correct and predicted labels is used as an evaluation criterion. The mathematical formula for calculating log loss is written as follows:

$$\text{Log Loss} = -\frac{1}{K} \sum_{i=1}^{K} \sum_{j=1}^{S} g_{ij} \, log \, (F_{ij}) \quad (1)$$

Here, *K* denotes the number of samples, and S denotes the number of classes. The true label of the class is represented by g, and the probability of the given sample is represented by *F*. The natural logarithm is used in the formula.

For the evaluation of our model, we used the following classification metrics: True Positive (TP), True Negative (TN), False Positive (FP), and False Negative (FN). We calculate Recall, True Positive Rate (TPR), False Positive Rate (FPR), Precision, Specificity, Sensitivity, F1 score, and Accuracy using these measures.

***Recall***: It represents a model's ability to find all relevant cases within a dataset. Mathematically, Recall is defined as the number of true positives divided by the total number of true positives plus the number of false negatives.

$$\text{Recall} = \text{TPR} = \frac{TP}{TP+FN} \quad (2)$$

***Precision***: It indicates a classification model's ability to identify only relevant data points. Precision is defined as the number of true positives divided by the number of true positives plus the number of false positives.

$$\text{Precision} = \frac{TP}{TP+FP} \quad (3)$$

***Specificity*** refers to the number of correctly predicted negative records. It helps determine how well our model predicts the class that we want to label as the negative class. In some ways, it is like Recall for the negative class.

$$\text{Specificity} = \frac{TN}{TN+FP} \quad (4)$$

***Sensitivity*** refers to the number of positive records correctly predicted. For the class that we want to declare as the positive class, sensitivity is the same as Recall.

$$\text{Sensitivity} = \frac{TP}{TP+FN} \quad (5)$$

Additionally, ***accuracy*** is another evaluation metric used to assess the performance of our model. It is mathematically defined as follows:

$$\text{Accuracy} = \frac{Tp+Tn}{Tp+TN+FP+FN} \quad (6)$$

The ***F1-score*** is the harmonic mean of precision and Recall, using the following equation to account for both metrics:

$$\text{F1– score} = 2*(\frac{Precision*Recall}{Precision + Recall}) \quad (7)$$

## 6  Database for Chest X-rays

A collaborative team of researchers from Qatar University, the University of Dhaka in Bangladesh, along with their partners from Pakistan and Malaysia, collaborated with medical professionals to develop a comprehensive database of chest X-ray images. This database includes cases of COVID-19-positive individuals, as well as normal and viral pneumonitis cases. The dataset is being made available in stages, with the initial release comprising 219 COVID-19, 1341 normal, and 1345 viral pneumonia chest X-ray (CXR) images. Subsequently, the COVID-19 class was expanded to include 1200 CXR images in the first update. In the second update, the database was further expanded to encompass 1345 viral pneumonia cases, 3616 COVID-19 positive cases, 10,192 normal cases, 6012 instances of lung opacity (non-COVID lung infection), and an additional 6012 COVID-19 positive cases. For our research, we are specifically interested in two datasets: one containing 10,192 normal chest X-rays and another containing 3616 COVID-19 positive chest X-rays [21].

The dataset curated by Chowdhury et al. [22] consists of 13,808 chest images captured in the Posterior-Anterior (PA) or anterior-posterior (AP) view. Each sample in the database has a resolution of 1024×1024 pixels and is stored in Portable Network Graphics (PNG) format. To facilitate compatibility with popular Convolutional Neural Networks (CNNs), the images have been resized to standard dimensions of 224×224 or 227×227 .

In this study, the dataset of 13,808 images was divided into training, validation, and test sets with a ratio of 80:10:10. The corresponding counts for each dataset are as follows:

1. **Training Dataset**: This dataset comprises a total of 13,808 images, with 2894 being COVID-19 positive chest X-rays and 8154 being normal chest X-rays.
2. **Validation Dataset**: The validation dataset consists of 1381 images, including 361 COVID-19 positive chest X-rays and 1019 normal chest X-rays.
3. **Testing Dataset**: The testing dataset contains 1381 images, with 361 COVID-19 positive chest X-rays and 1019 normal chest X-rays.

Table 1 provides an overview of the distribution of images across the training, validation, and test sets.

Table 1. Distribution of chest X-ray images after splitting

| Classes | Training Set | Validation Set | Testing Set | Total |
|---|---|---|---|---|
| **COVID-19** | 2894 | 361 | 361 | 3616 |
| **NORMAL** | 8154 | 1019 | 1019 | 10192 |

In the dataset, normal cases equal (0), and COVID-19 positive cases equal (1), as shown in Fig. 4.

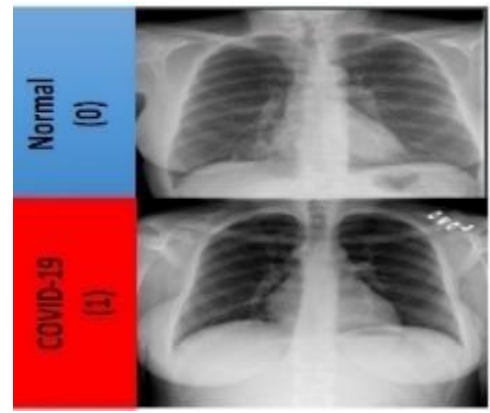

**Fig. 4**. The dataset of normal cases and COVID-19-positive cases

# 7 Experimental Results

The experiments were conducted on an HP EliteBook operating on Windows 10 Pro 64-bit. The system specifications included a 3.8 GHz Core i7 processor, 16 GB of RAM, an Intel HD Graphics 4600 GPU, and 1 TB of storage. The experiments were implemented using Python 3.9 with the Keras and TensorFlow libraries within the PyCharm program.

The results of our proposed framework showcased exceptional performance in classifying COVID-19 and normal cases, achieving an accuracy of 99.88% and a validation rate of 96.50%. A summary of the classification results is presented in Table 2.

Table 2: Classification results

| Metrics | Precisiontab | Recall | F1-score | Accuracy |
|---|---|---|---|---|
| Training | 100% | 100% | 100% | 99.88% |
| validation | 98.88% | 98% | 98.45% | 96.50% |

According to Table 2, our proposed approach achieves the highest accuracy of 99.88% in classifying COVID-19 and normal cases. It is accompanied by a validation rate of 96.5%, precision and recall rates of 100%, and an F1 score of 100% based on the training rate. However, for the validation rate, the precision and recall rates are 98%, and the F1 score is 98.45% with a precision of 98.88% and recall rate of 98%.

Fig. 5 provides a visual representation comparing the pooling of a 3-by-3 image with a stride of 2 and the pooling of a 2-by-2 image with a stride of 3. The illustration concludes that while reducing the image features, minimal information loss occurs.

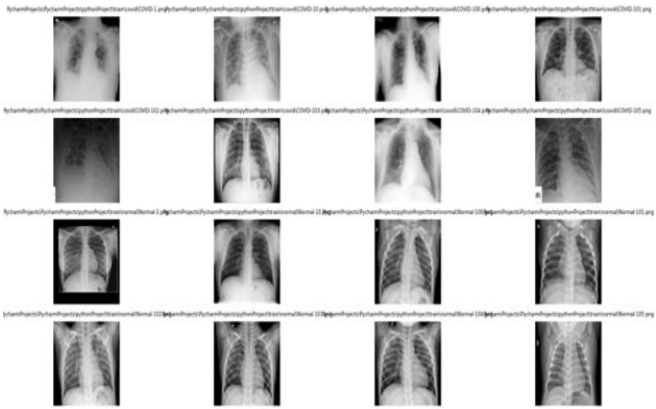

**Fig. 5**. Effect of stride and pooling on image resolution

Before the completely linked CNN layer, feature maps from the three sequential layers are concatenated. Weights are calculated using the Glorot technique [23], the Adam optimizer [24], and a learning rate of 0.001, 400 epochs, and 32 mini-batch sizes. Table 3 depicts the Proposed model based on CNN's structure.

Table 3: Proposed model based on CNN

| Layer (type) | Output Shape | Parameter |
|---|---|---|
| Conv2d (Conv2D) | (None, 200, 200, 64) | 6,976 |
| Max_pooling2d(Maxpooling2D) | (None, 100, 100, 64) | 0 |
| Dropout (Dropout) | (None, 100, 100, 64) | 0 |
| Conv2d_1(Conv2D) | (None, 100, 100, 128) | 295,040 |
| Max_pooling2d_1( Maxpooling2D) | (None, 50, 50, 128) | 0 |
| Dropout_1 (Dropout) | (None, 50, 50, 128) | 0 |
| Conv2d_2(Conv2D) | (None, 50, 50, 256) | 1,179,904 |
| Max_pooling2d_2( Maxpooling2D) | (None, 25, 25, 256) | 0 |
| Dropout_2 (Dropout) | (None, 25, 25, 256) | 0 |
| Flatten (Flatten) | (None, 160000) | 0 |
| Dense (Dense) | (None, 512) | 81,920,512 |
| Dropout_3 (Dropout) | (None, 512) | 0 |
| Dense_1 (Dense) | (None, 1) | 513 |

# 8 Evaluation of Two and Three Layers of the CNN Model with different Epochs

To validate the output of the presented approach, we implemented and tested the approach using two different numbers of epochs (150 and 400) with two and three layers of the CNN model. A series of experiments were conducted to assess the performance of the proposed model. The results encompass various performance metrics, including accuracy, precision, recall, and F1-score.

### A. Using 150 Epochs

Our proposed model, Deep-COVID-19, based on CNN, underwent training for 150 epochs with early stopping. The testing phase of Deep-COVID-19 was performed on the remaining 20% of the dataset after training all models on 80% of the data. Figure 6 illustrates the relationship between the number of epochs and the loss value. In the initial epoch, the training and validation losses for the two-layer CNN model were 3.0667 and 2.0134, respectively, while for the three-layer CNN model, they were 0.4007 and 0.3711, respectively. However, by the fifth epoch, the training and validation losses for the two-layer model significantly dropped to 0.2554 and 0.3139, respectively, and for the three-layer model, they dropped to 0.2439 and 0.2941, respectively. Subsequently, the training loss for the two-layer model gradually decreased to 0.0179 at the 120th epoch, and for the three-layer model, it gradually decreased to 0.036 at the 141st epoch. Similarly, the validation losses for the two-layer model gradually decreased to 0.189 at the 37th epoch, and for the three-layer model, they gradually decreased to 0.1098 at the 121st epoch.

Fig. 7 depicts the relationship between the number of epochs and the accuracy value. In the first epoch, the training and validation accuracies for the two-layer CNN model were 0.74 and 0.625, respectively, while for the three-layer CNN model, they were 0.855 and 0.855, respectively. However, by the fifth epoch, the training and validation accuracies for the two-layer model dramatically increased to 0.9087 and 0.9038, respectively, and for the three-layer model, they increased to 0.905 and 0.875, respectively. Eventually, the training accuracy for the two-layer model gradually increased to 0.9962 at the 120th epoch, and for the three-layer model, it gradually increased to 0.9975 at the 133rd epoch. Similarly, the validation accuracy for the two-layer model gradually increased to 0.96 at the 125th epoch, and for the three-layer model, it gradually increased to 0.955 at the 106th epoch.

Fig. 8 presents the relationship between the number of epochs and the precision value. In the first epoch, the training and validation precisions for the two-layer CNN model were 0.75 and 0.6374, respectively, while for the three-layer CNN model, they were 0.866 and 0.8901, respectively. However, by the fifth epoch, the training and validation precisions for the two-layer model significantly increased to 0.9383 and 0.933, respectively. For the three-layer model, they increased to 0.9355 and 0.9747, respectively. Finally, the training precision for the two-layer model gradually increased to 1 at the 120th epoch, and for the three-layer model, it gradually increased to 1 at the 133rd epoch. Similarly, the validation precision for the two-layer model gradually increased to 0.9759 at the 49th epoch, and for the three-layer model, it gradually increased to 0.9778 at the 24th epoch.

Fig. 9 demonstrates the relationship between the number of epochs and the recall value. In the first epoch, the training and validation recalls for the two-layer CNN model were 0.72 and 0.58, respectively, while for the three-layer CNN model, they were 0.84 and 0.81, respectively. However, by the fifth epoch, the training and validation recalls for the two-layer model significantly increased to 0.875 and 0.87, respectively, and for the three-layer model, they increased to 0.87 and 0.94, respectively. Eventually, the training recall for the two-layer model gradually increased to 0.9925 at the 120th epoch, and for the three-layer model, it gradually increased to 0.9935 at the 133rd epoch. Similarly, the validation recall for the two-layer model gradually increased to 0.99 at the 112th epoch, and for the three-layer model, it gradually increased to 0.98 at the third epoch.

Fig. 10 shows the relationship between the number of epochs and the F1 score value. In the first epoch, the training and validation F1 scores for the two-layer CNN model were 2.88 and 2.32, respectively, and for the three-layer CNN model, they were 3.36 and 3.24, respectively. In contrast, both models' training and validation F1 scores dramatically increased to 3.5 and 3.48, respectively, at the fifth epoch.

Finally, the training F1 score for the two-layer CNN model gradually increased to 3.97 at 120 epochs, and the training F1 score for the three-layer CNN model gradually increased to 3.974 at 133 epochs. The validation F1 score for the two-layer CNN model gradually increased to 3.96 at 122 epochs, and the validation F1 score for the three-layer CNN model gradually increased to 3.92 at 133 epochs.

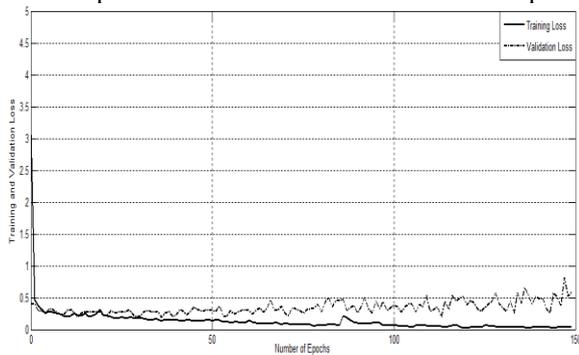

Loss 150 using 2 layers

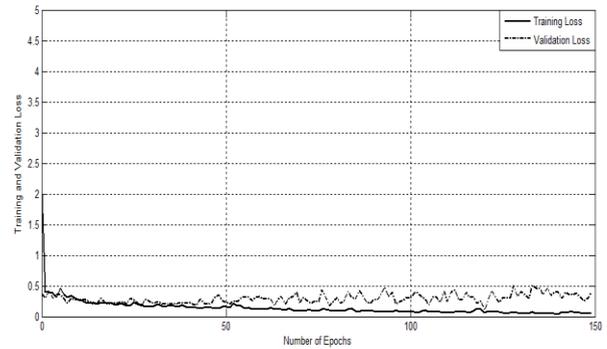

Loss 150 using 3 layers

**Fig. 6.** Model loss of our proposed model based on CNN using (two & three) layers

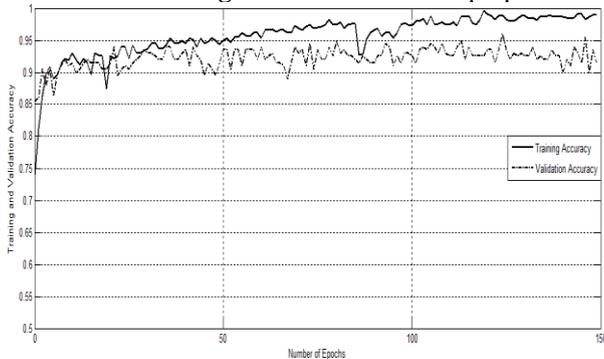

accuracy 150 using 2 layers

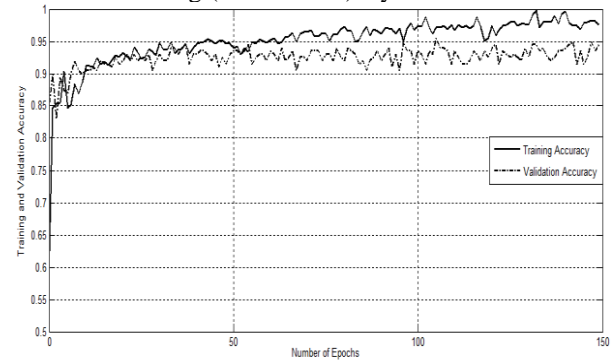

accuracy 150 using 3 layers

**Fig. 7**. Model accuracy of our proposed model based on CNN using (two & three) layers

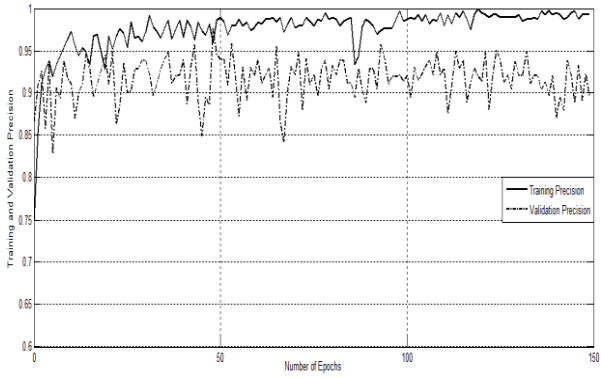
precision 150 using 2 layers

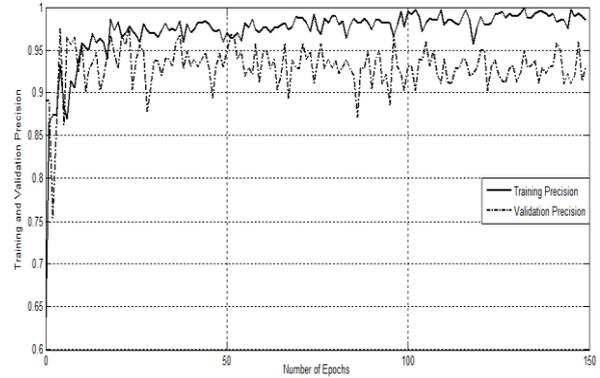
precision 150 using 3 layers

**Fig 8**. model precision of our Proposed model based on CNN using (two & three) layers

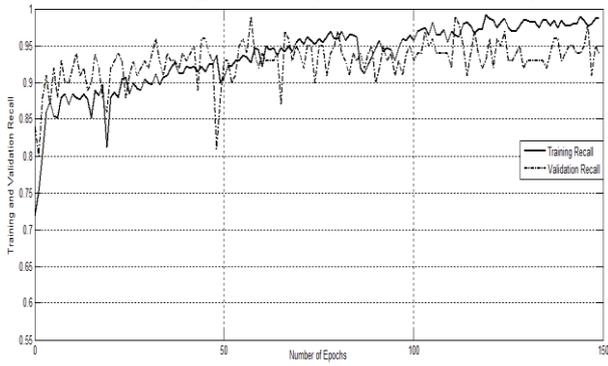
recall 150 using 2 layers

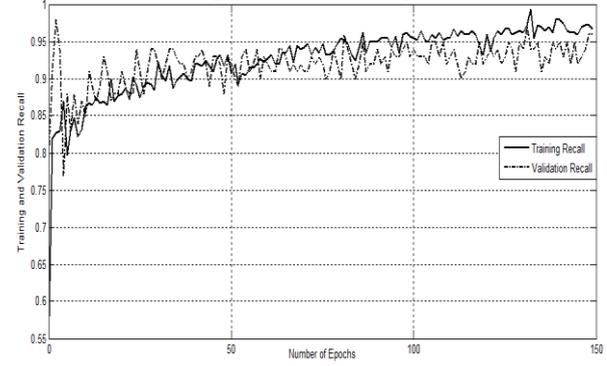
recall 150 using 3 layers

**Fig. 9.** model recall of our Proposed model based on CNN using (two & three) layers

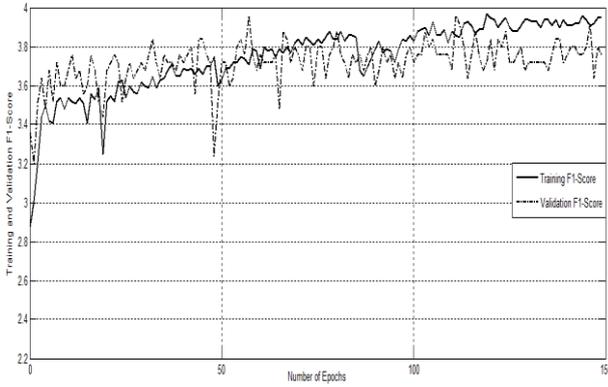
F1-Score 150 using 2 layers

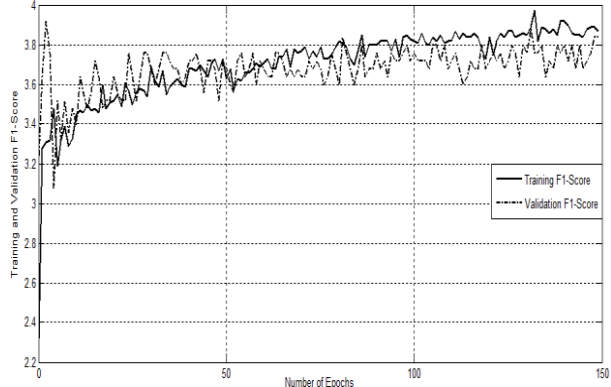
F1-Score 150 using 3 layers

**Fig. 10.** Model F1-Score of our Proposed model based on CNN using (two & three) layers

### B. Using 400 epochs

Deep-COVID-19, our proposed model based on CNN, was trained for 400 epochs with early stopping. It was tested on the remaining 20% of the dataset after all models were trained on 80% of it. Fig. 11 shows the relationship between the number of epochs and the loss value. At the first epoch, the loss values for the training and validation sets are 1.8984 and 1.7683 for the two-layer model and 0.3852 and 0.4419 for the three-layer model. The loss values for both models decrease dramatically to 0.3394 and 0.3229 at the fifth epoch. The training loss for the two-layer model gradually decreases to 0.003 at 330 epochs and the training loss for the three-layer model gradually decreases to 0.0011 at 361 epochs. The validation losses for the two-layer model gradually decrease to 0.19 at 70 epochs and the validation losses for the three-layer model gradually decrease to 0.1418 at 75 epochs.

Fig. 12 shows the relationship between the number of epochs and the accuracy value. At the first epoch, the accuracy values for the training and validation sets are 0.665 and 0.6388 for the two-layer model and 0.845 and 0.785 for the three-layer model. The accuracy values for both models increase dramatically to 0.8775 and 0.9013 at the fifth epoch. The training accuracy for the two-layer model gradually increases to 0.9987 at 258 epochs and the training accuracy for the three-layer model gradually increases to 0.9988 at 373 epochs. The validation accuracy for the two-layer model gradually increases to 0.955 at 318 epochs and the validation accuracy for the three-layer model gradually increases to 0.965 at 159 epochs.

Fig. 13 shows the relationship between the number of epochs and the precision value. At the first epoch, the precision values for the training and validation sets are 0.6658 and 0.6907 for the two-layer model and 0.8485 and 0.728 for the three-layer model. The precision values for both models increase dramatically to 0.9081 and 0.9303 at the fifth epoch. The training precision for the two-layer model gradually increases to 1 at 196 epochs, and the training precision for the three-layer model gradually increases to 1 at 188 epochs. The validation precision for the two-layer model gradually increases to 0.9892 at 247 epochs, and the validation precision for the three-layer model gradually increases to 0.9888 at 324 epochs.

Fig.14 shows the relationship between the number of epochs and the recall value. At the first epoch, the recall values for the training and validation sets are 0.6625 and 0.5025 for the two-layer model and 0.84 and 0.91 for the three-layer model. The recall values for both models increase dramatically to 0.84 and 0.8675 at the fifth epoch. The training recall for the two-layer model gradually increases to 1 at 333 epochs, and the training recall for the three-layer model gradually increases to 1 at 204 epochs. The validation recall for the two-layer model gradually increases to 0.97 at 317 epochs and the validation recall for the three-layer model gradually increases to 0.98 at 360 epochs.

Fig. 15 shows the relationship between the number of epochs and the F1 score value for the two-layer and three-layer CNN models. At the first epoch, the F1 scores for the training and validation sets are 2.65 and 2.01 for the two-layer model and 3.36 and 3.64 for the three-layer model. The F1 scores for both models increase dramatically to 3.36 and 3.47 at the fifth epoch. The training F1 score for the two-layer model gradually increases to 4 at 330 epochs, and the training F1 score for the three-layer model gradually increases to 4 at 204 epochs. The validation F1 score for the two-layer model gradually increases to 3.88 at 317 epochs and the validation F1 score for the three-layer model gradually increases to 3.92 at 360 epochs.

By comparing the results of our proposed model based on CNN with other models in the literature, it was clear that our model has higher accuracy in classifying COVID-19 and normal cases. Our model achieved an accuracy of 99.88%, a validation rate of 96.5%, a precision of 100%, a recall of 100%, and an F1 score of 100%.

### C. Comparison of Deep Learning Models for the Diagnosis of COVID-19

This section provides a systematic review of research studies on the use of deep Learning in the diagnosis of COVID-19. This process uses images from X-rays and CT scans.

Several research papers have focused on the use of deep-learning techniques to diagnose COVID-19 patients. Table 4 shows various deep-learning methods for diagnosing COVID-19 using X-ray images. The study in [25] used datasets from Kaggle. The experiments used X-ray images from 50 patients (abnormal) and 50 normal people.

The study [25] obtained a classification accuracy value of 87% using the Inception-ResNetV2 model. Wang et al. [26] achieved an accuracy value of 92.6% using the COVID-Net network architecture. Ghoshal et al. [27] achieved an accuracy value of 92.9% using the Bayesian ResNet50V2 model.

Loey et al. [28] used six different deep learning models: AlexNet for four classes, AlexNet for three classes, GoogleNet for four classes, GoogleNet for three classes, ResNet18 for four classes, and ResNet18 for four classes. These models achieved classification accuracy values of 66.67%, 85.19%, 80.56%, 81.48%, 69.46%, and 81.48%, respectively.

Ozturk et al. [29] achieved an accuracy value of 87.02% using the DarkCovidNet model (multi-class classification task). Mahmud et al. [30] achieved an accuracy value of 92.1% using the Residual Network model (COVID-19/Normal). Khan et al. [31] achieved an accuracy value of 89.6% using the CoroNet model. Waheed et al. [32] achieved an accuracy value of 85% using the Actual data (CNN-AD) model. In this paper, our proposed model based on CNN achieved the highest accuracy, reaching 99.88%.

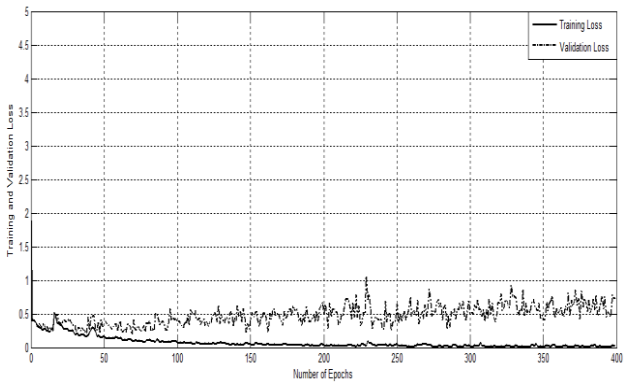
Loss 400 using 2 layers

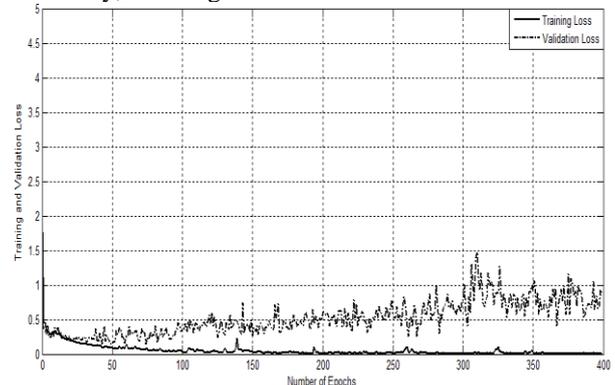
Loss 400 using 3 layers

**Fig. 11.** model loss of our Proposed model based on CNN using (two & three) layers

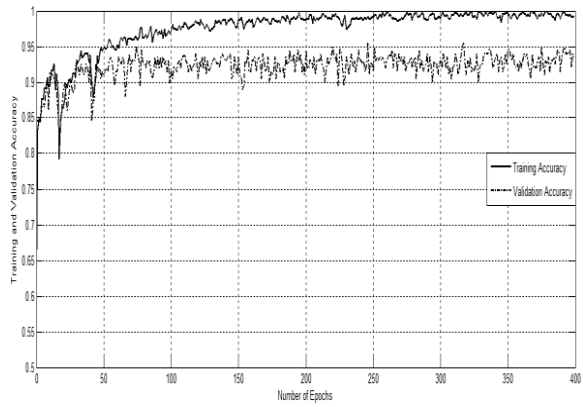 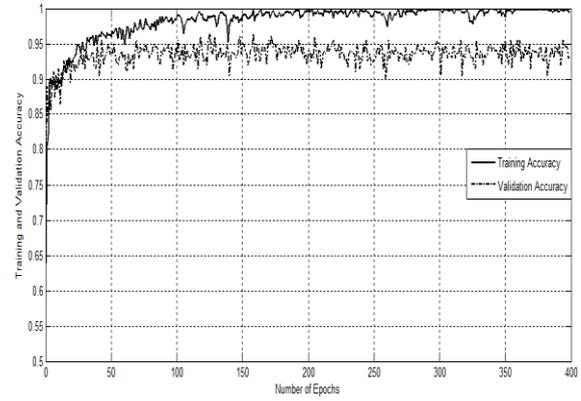

accuracy 400 using 2 layers          accuracy 400 using 3 layers

**Fig 12.** model accuracy of our Proposed model based on CNN using (two & three) layers

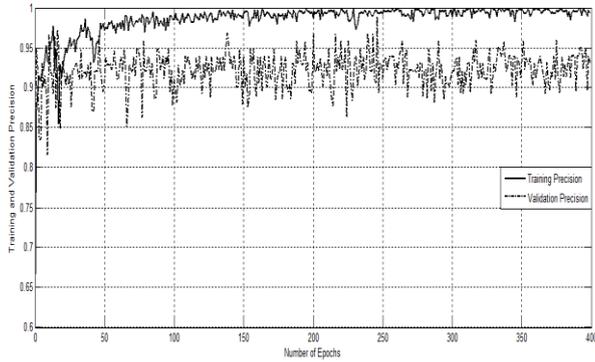 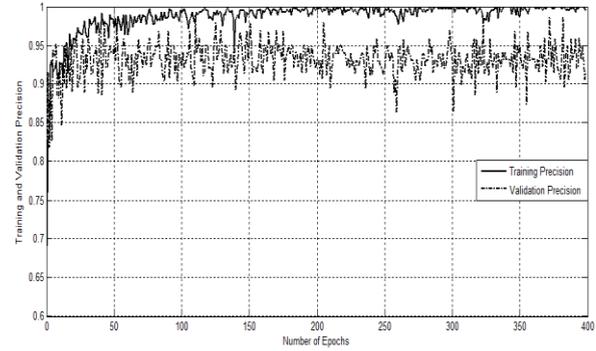

precision 400 using 2 layers          precision 400 using 3 layers

**Fig. 13.** Model precision of our proposed model based on CNN using (two & three) layers

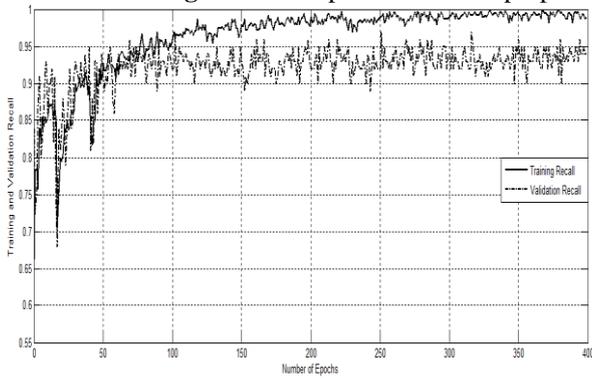 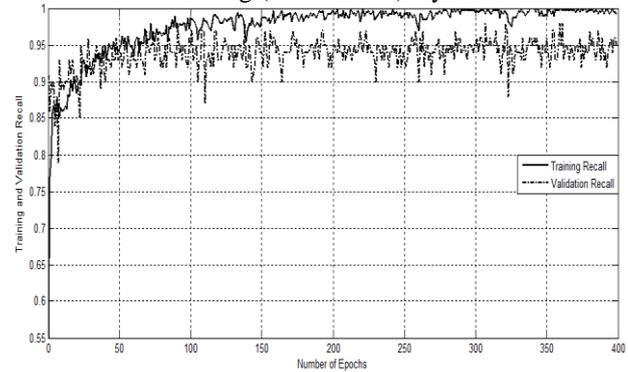

recall 400 using 2 layers          recall 400 using 3 layers

**Fig. 14.** model recall of our Proposed model based on CNN using (two & three) layers

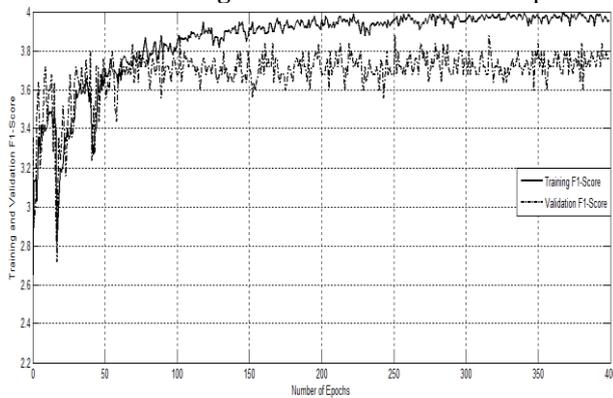 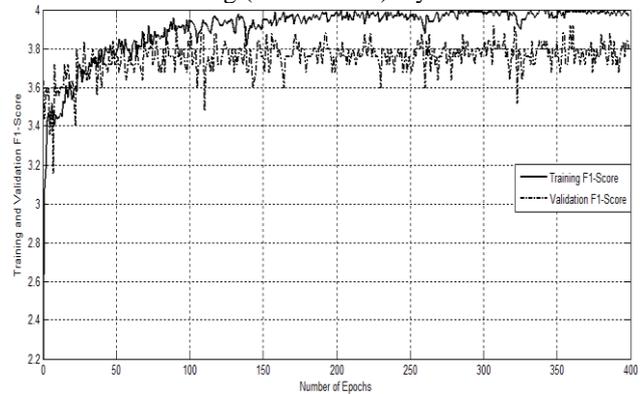

F1-Score 400 using 2 layers          F1-Score 400 using 3 layers

**Fig. 15.** Model F1-Score of our Proposed model based on CNN using (two & three) layers.

Table 4: Deep Learning for COVID-19 diagnosis in X-ray images.

| Refs. | Methods | Accuracy |
|---|---|---|
| [47] | AlexNet for 4 classes | 66.67% |
| | ResNet18 for 4 classes | 69.46% |
| | GoogleNet for 4 classes | 80.56% |
| | GoogleNet for 3 classes | 81.48% |
| | ResNet18 for 3 classes | 81.48% |
| [51] | Actual data (CNN-AD) | 85% |
| [47] | AlexNet for 3 classes | 85.19% |
| [44] | Inception ResNetV2 | 87% |
| [48] | DarkCovidNet (multi-class classification task) | 87.02% |
| [50] | CoroNet | 89.6% |
| [49] | Residual Network (COVID-19/ Normal) | 92.1% |
| [45] | COVID-Net network architecture | 92.6% |
| [46] | Bayesian ResNet50V2 | 92.9% |
| Proposed model based on CNN | Modified convolutional neural network (CNN) | 99.88% |

We also compared our proposed model based on CNN with three pre-trained models: VGG16, VGG19, and MobileNet. We used the same dataset for all models.

Transfer learning is a common practice in computer vision. A pre-trained model is one that has been trained on a large benchmark dataset to solve a problem similar to the one we want to solve. Because of the computational cost of training such models, it is common practice to import and use models from published literature (e.g., VGG16, VGG19, MobileNet).

Transfer learning is a powerful technique in computer vision because it allows us to build accurate models in a timely manner. With transfer learning, we do not start from scratch; instead, we start from patterns that have been established when solving a different problem. This allows us to build on existing knowledge rather than starting from scratch [33].

Fig. 16 shows the relationship between the number of epochs and the accuracy value for the three pre-trained models. At the first epoch, the training and validation accuracy for all three models are low. However, the accuracy for all models gradually increases over time. At 400 epochs, the training accuracy for VGG16, VGG19, and MobileNet are 0.925, 0.946, and 0.957, respectively. The validation accuracy for VGG16, VGG19, and MobileNet are 0.895, 0.927, and 0.935, respectively.

Therefore, the proposed model outperforms the three pre-trained models. Our model achieved a higher accuracy at 400 epochs than the three pre-trained models. Due to the limited scope of this paper, we have summarized the results of the comparison in the table 5.

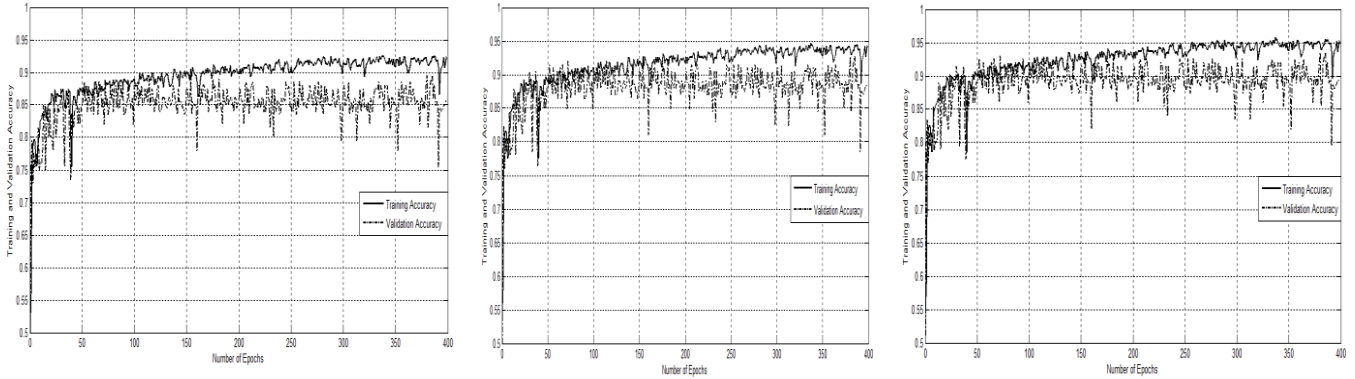

accuracy 400 using VGG16     accuracy 400 using VGG19     accuracy 400 using MobileNet

**Fig. 16**. Model accuracy of three pre-trained models (VGG16, VGG19 and MobileNet)

Table 5: Classification results

| Models | Dataset | Precision | Recall | F1-score | Accuracy |
|---|---|---|---|---|---|
| VGG16 | Train | 71% | 70% | 70.5% | 92.5% |
| | Valid | 68.9% | 67% | 67.9% | 89.5% |
| VGG19 | Train | 82% | 82% | 82% | 94.6% |
| | Valid | 79.3% | 78% | 78.6% | 92.7% |
| MobileNet | Train | 93% | 93% | 93% | 95.7% |
| | Valid | 89.6% | 87% | 88.3% | 93.5% |
| Proposed model based on CNN | Train | 100% | 100% | 100% | 99.88% |
| | Valid | 98.88% | 98% | 98.45% | 96.50% |

## 9 Conclusion and Future Work

This paper presents a framework based on an IoT fog-cloud computing architecture for identifying COVID-19. We also propose a model based on CNN that is deployed and implemented on a fog computing layer to detect COVID-19 from chest X-ray (CXR) images. We evaluate the performance of the proposed model by studying its categorization accuracy. The proposed model was experimented with using two layers of CNN, and the results showed that the training and validation accuracy gradually increased to 99.87% and 95.50%, respectively.

The proposed model was also experimented with three layers of CNN. The results showed that the training and validation accuracy increased to 99.88% and 96.50%, respectively. The proposed model was then compared with other studies and with three pre-trained models: VGG16, VGG19, and MobileNet. The results showed that the accuracy of the proposed model was higher than the other models. The proposed model achieved an accuracy of 99.88% in classifying COVID-19 and normal cases, along with a validation rate of 96.5%, precision of 100%, recall of 100%, and F1 score of 100%. In the future, we plan to:

- Secure user multimedia data in the cloud using fog computing.
- Focus on authentication and key agreement using different authentication algorithms.
- Use electrocardiography (ECG) to detect COVID-19, as recent forms of COVID-19 can affect the cardiovascular system.
- Investigate the use of empirical wavelet transform (EWT) and principal component analysis (PCA) for data filtering.

## Acknowledgment

This research paper is not supported by any company or organization; it relies solely on the efforts of the authors.

## Funding

Not applicable.

## Conflict of interests

The authors declare that they have no conflict of interest.

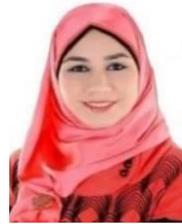

Alzahraa Elsayed received the MSc in Communications engineering and Computers Engineering from the University of Al Azhar, Egypt (2018), where she is currently pursuing the PhD in Communications engineering from 2019 to 2022, her research interests include fog computing, cloud computing, and internet of things (IOT) technologies.

E-mail: alzahraa.salah@azhar.edu.eg

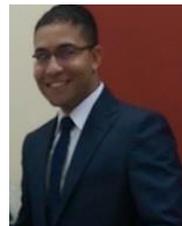

Khalil Mohamed received a Ph.D. in robotics and control engineering from Al-Azhar University, Egypt in 2019. He is currently an assistant professor at Systems and Computers Engineering Department, at Al-Azhar University, Egypt.
His research interests include AI, Machine learning, Deep learning, Reinforcement learning, Robotics, Control theory, Intelligent Control Systems, Automotive Control Systems, Robust Control, Stochastic Control, Motion and Navigation Control, Traffic and Transport Control, Predictive control, Optimal control, Mathematics, Optimization, Task assignment in multi-robot systems, Task decomposition.

E-mail: eng.khalil@azhar.edu.eg

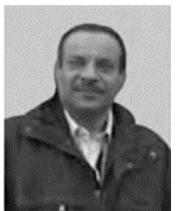

Hany Harb received the B.Sc. degree in computers and control engineering from Faculty of Engineering, Ain Shams University, Egypt in 1978, the M.Sc. degree in computers and systems engineering from Faculty of Engineering, Al-Azhar University, Egypt in 1981. He also received the Ph.D. degree in computer science and the M.Sc. degree in operations research (MSOR) from Institute of Technology (IIT), USA in 1986 and 1987, respectively. He is a professor of software engineering in System Engineering Department, Faculty of Engineering, Al Azhar University, Egypt. His research interests include artificial intelligence, cloud computing, and distributed systems.

E-mail: harbhany@yahoo.com